\pdfoutput=1

\documentclass[11pt]{article}

\usepackage{acl}
\usepackage{amsmath} 
\usepackage{times}
\usepackage{latexsym}
\usepackage{adjustbox}
\usepackage[T1]{fontenc}

\usepackage[utf8]{inputenc}

\usepackage{microtype}
\usepackage{graphicx}

\title{VisualRWKV-HD and UHD: Advancing High-Resolution Processing for Visual Language Models}

\author{
Zihang Li$^{1,2}$, Haowen Hou$^2$\thanks{Corresponding author.}\\
$^1$Department of Artificial Intelligence, Shenzhen University, Shenzhen, China\\
$^2$Guangdong Laboratory of Artificial Intelligence and Digital Economy (SZ), Shenzhen, China\\
\texttt{2410815011@mails.szu.edu.cn, houhaowen@gml.ac.cn}
}

\begin{document}
\maketitle
\begin{abstract}
Accurately understanding complex visual information is crucial for visual language models (VLMs). Enhancing image resolution can improve visual perception capabilities, not only reducing hallucinations but also boosting performance in tasks that demand high resolution, such as text-rich or document analysis.
In this paper, we present VisualRWKV-HD and VisualRWKV-UHD, two advancements in the VisualRWKV model family, specifically designed to process high-resolution visual inputs.
For VisualRWKV-HD, we developed a lossless downsampling method to effectively integrate a high-resolution vision encoder with low-resolution encoders, without extending the input sequence length.
For the VisualRWKV-UHD model, we enhanced image representation by dividing the image into four segments, which are then recombined with the original image. This technique allows the model to incorporate both high-resolution and low-resolution features, effectively balancing coarse and fine-grained information. As a result, the model supports resolutions up to 4096 x 4096 pixels, offering a more detailed and comprehensive visual processing capability.
Both VisualRWKV-HD and VisualRWKV-UHD not only achieve strong results on VLM benchmarks but also show marked improvements in performance for text-rich tasks.
\end{abstract}

\section{Introduction}
\begin{figure*}
    \centering
    \includegraphics[width=1\linewidth]{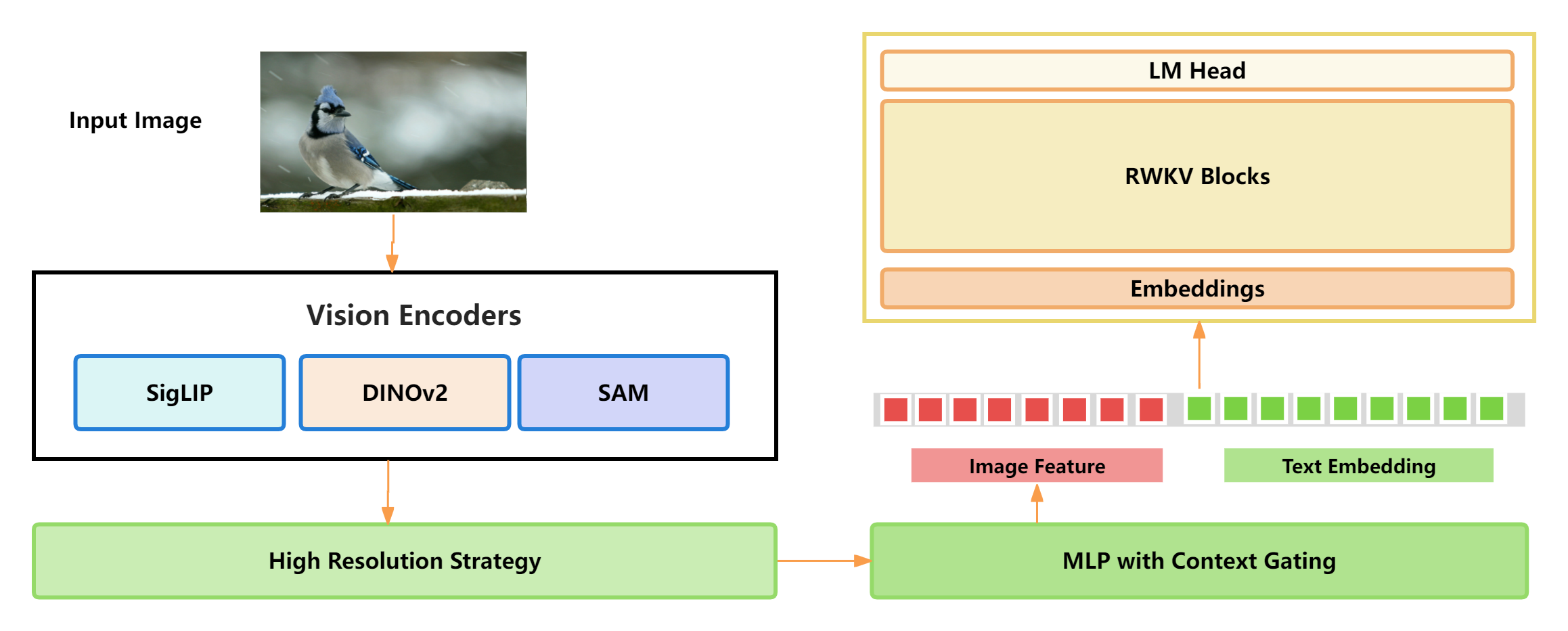}
    \caption{VisualRWKV-HD and UHD Overview. The input image is processed through three vision encoders and a high-resolution strategy, followed by a multi-layer perceptron (MLP) with context gating to generate image features.}
    \label{fig:enter-label}
\end{figure*}
With the recent significant advancements in large language models\cite{akyürek2024incontext}(LLMs)\cite{achiam2023gpt}, visual language models (VLMs) have also rapidly progressed.The efforts\cite{peng2023rwkv} to extend LLMs to handle visual inputs through visual instruction tuning are growing steadily\cite{liu2024improved}. Concurrently, visual language models with linear time complexity, such as VisualRWKV\cite{hou2024visualrwkv} and VL-Mamba\cite{qiao2024vl}, have also been proposed. However, research on efficiently processing high-resolution visual inputs within linear visual language models remains lacking.
Enhancing image resolution can improve visual perception capabilities, not only reducing hallucinations but also boosting performance in tasks that demand high resolution, such as text-rich or document analysis. However, the challenge with high-resolution images is that they often lead to increased computational demands and longer input sequences, which can hinder model efficiency and performance\cite{hou2024visualrwkv}.To address these challenges, in this paper, we introduce two novel advancements in the VisualRWKV model family: VisualRWKV-HD and VisualRWKV-UHD. Figure 1 shows the overview of VisualRWKV-HD and UHD. These models are specifically designed to capitalize on the benefits of high-resolution visual inputs while maintaining computational efficiency and performance, marking the first advancement of linear RNN models to solve high-resolution tasks. There are four main contributions in this paper:
\begin{enumerate}
\item VisualRWKV-HD with Ensemble of Encoders: We introduce an ensemble of encoders, where the input image size is fixed at 1024 during the pre-training of SigLip, DINOv2, and Segment Anything Model (SAM). As a base model, SAM is expected to exhibit generalization capabilities across various downstream tasks with different image sizes. This is particularly important for high-definition (HD) datasets, which have larger sizes and more detail. SAM performs well when the image resolution aligns with its training resolution of 1024. Therefore, we used SAM to support resolutions up to 1024 x 1024, achieving significant performance improvements on several benchmarks, such as TextVQA. 

\item VisualRWKV-UHD: In VisualRWKV-UHD, the image is divided into four segments and then recombined, allowing the image features to contain both high-resolution and low-resolution information. This approach balances coarse and fine-grained features, allowing the model to support resolutions up to 4096 x 4096 while ensuring that the number of image tokens does not exceed 1024.

\item MLP with Context Gating: We found that excessive feature information led to internal competition within the model. To address this, we introduced MLP with Context Gating to replace the linear projection layer, stabilizing the training process and improving performance.

\item High-Resolution and Low-Resolution Alignment: To align high-resolution vision encoders with low-resolution modules, we propose an alternative approach that combines every 2x2 block (each containing four adjacent vectors) into a new channel dimension. This method preserves information without the need for additional training.
\end{enumerate}
In summary, this study presents the VisualRWKV-HD and VisualRWKV-UHD models. VisualRWKV-HD introduces a pre-trained ensemble of vision encoders, increasing the supported resolution to 1024 x 1024. On the other hand, VisualRWKV-UHD is designed for efficiency, capable of handling even higher resolutions with support for inputs up to 4096 x 4096 pixels. Importantly, it maintains a maximum of 1024 image tokens, whereas tiled approaches tend to use an excessive number of image tokens, resulting in longer processing times. Comprehensive experiments on eight popular benchmarks demonstrate the effectiveness of these models, particularly in tasks that require high-resolution visual processing. In addition, an in-depth analysis is provided to offer a deeper understanding of the model's improvements and capabilities.
\section{Related Works}
\subsection{Linear Visual Language Models}
In the rapidly evolving field of visual language models (VLMs)\cite{bai2023qwen}, linear models have introduced innovative strategies for integrating image and text data. These models, characterized by their straightforward architectures, have proven effective in tasks such as image captioning and visual question answering.
Linear Visual Language Models have made significant progress in efficiently merging visual and textual information. VL-Mamba\cite{qiao2024vl} stands out for its emphasis on robust alignment between visual and textual modalities, leveraging a modular design that enables flexibility across various applications. This adaptability allows the model to cater to a wide range of tasks while maintaining high performance.On the other hand, VisualRWKV\cite{hou2024visualrwkv}, including its high-definition variants, tackles the complexities of high-resolution image processing by employing techniques such as lossless down-sampling and image segmentation. These methods facilitate the effective integration of both high and low-resolution features, ensuring that essential details are preserved.
Together, these models not only signify important advancements in the VLM landscape but also underscore innovations in efficiency, detail retention, and multi-modal alignment. As they continue to evolve, these linear visual language models are set to propel further research, particularly in the domain of handling complex visual information.

\subsection{High-Resolution Visual Language Models}
LLaVA-UHD\cite{xu2024llava} is a model specifically designed for high-resolution visual tasks, integrating advanced multi-modal capabilities that enhance its applications in image captioning and visual question answering. One of its key strengths lies in its ability to seamlessly integrate visual and textual data, allowing it to generate contextually relevant outputs based on complex visual inputs. This integration is achieved through optimized model architecture, improving efficiency and adaptability across various tasks.
The model has been trained on extensive datasets, enhancing its generalization capabilities. This training enables LLaVA-UHD to effectively comprehend fine visual details and nuances, making it particularly suitable for applications that require high precision in visual understanding.

\subsection{Comparative Analysis}
To elucidate the innovations in VisualRWKV-HD and VisualRWKV-UHD, we compare them with existing models such as LLAVA-UHD and VisualRWKV. VisualRWKV combines visual encoders and language models for cross-modal understanding but struggles with high-resolution images due to computational bottlenecks. LLAVA-UHD addresses high-resolution processing by employing multi-scale feature fusion and adaptive resolution mechanisms, though this can lead to high computational demands. In contrast, VisualRWKV-HD uses a pre-trained SAM visual encoder to handle high-resolution inputs efficiently without increasing input sequence length, and it optimizes model architecture to reduce complexity. VisualRWKV-UHD further integrates high-resolution and low-resolution features through image segmentation and introduces the MLPWithContextGating mechanism for better detail retention and training stability. Overall, VisualRWKV-HD and VisualRWKV-UHD offer improved efficiency and detail preservation in high-resolution image processing compared to LLAVA-UHD and VisualRWKV.
\section{Method}
\subsection{VisualRWKV-HD}

\subsubsection{Ensemble of Encoders}

In previous versions of VisualRWKV, we utilized SIGLIP and DINO encoders focused on processing low-resolution images, achieving good results. In VisualRWKV-HD, we introduced a pre-trained high-resolution SAM vision encoder, enhancing the model's supported resolution to 1024 x 1024. This improvement significantly boosted the model's performance across multiple benchmark tests, making it more efficient and accurate in handling tasks that require rich details and visual complexity. The inclusion of the SAM encoder allows the model to better capture critical features in images, enhancing its overall visual understanding capabilities. 

\subsubsection{Lossless DownSampler}

SigLip and DINOv2, as encoders from the previous generation of VisualRWKV, have shown effectiveness in low-resolution image tasks. In this generation, VisualRWKV-HD introduces the SAM encoder. To address the alignment issue with the SAM encoder, we designed a lossless downsampler that combines 2x2 blocks (each containing four adjacent vectors) into a new channel dimension. This method allows the high-resolution vision encoder to effectively align with lower-resolution modules without losing information during training. Experimental results indicate that this approach effectively preserves image details and enhances model performance. 

You could use a formula to represent the process of combining the 2x2 blocks into a new channel dimension. Here’s a suggested formula:
\begin{equation}
    C_{new} = \text{Concat}(C_{1}, C_{2}, C_{3}, C_{4})
\end{equation}
\text{Where:}
\begin{itemize}
    \item $C_{new}$ represents the new channel dimension formed by concatenating the four blocks.
    \item $C_{1}, C_{2}, C_{3}, C_{4}$ are the 2x2 blocks, each containing four adjacent vectors.
\end{itemize}
This formula illustrates how the new channel dimension is created by combining the lower-resolution representations effectively.

\subsection{VisualRWKV-UHD}

\begin{figure*}
    \centering
    \includegraphics[width=1\linewidth]{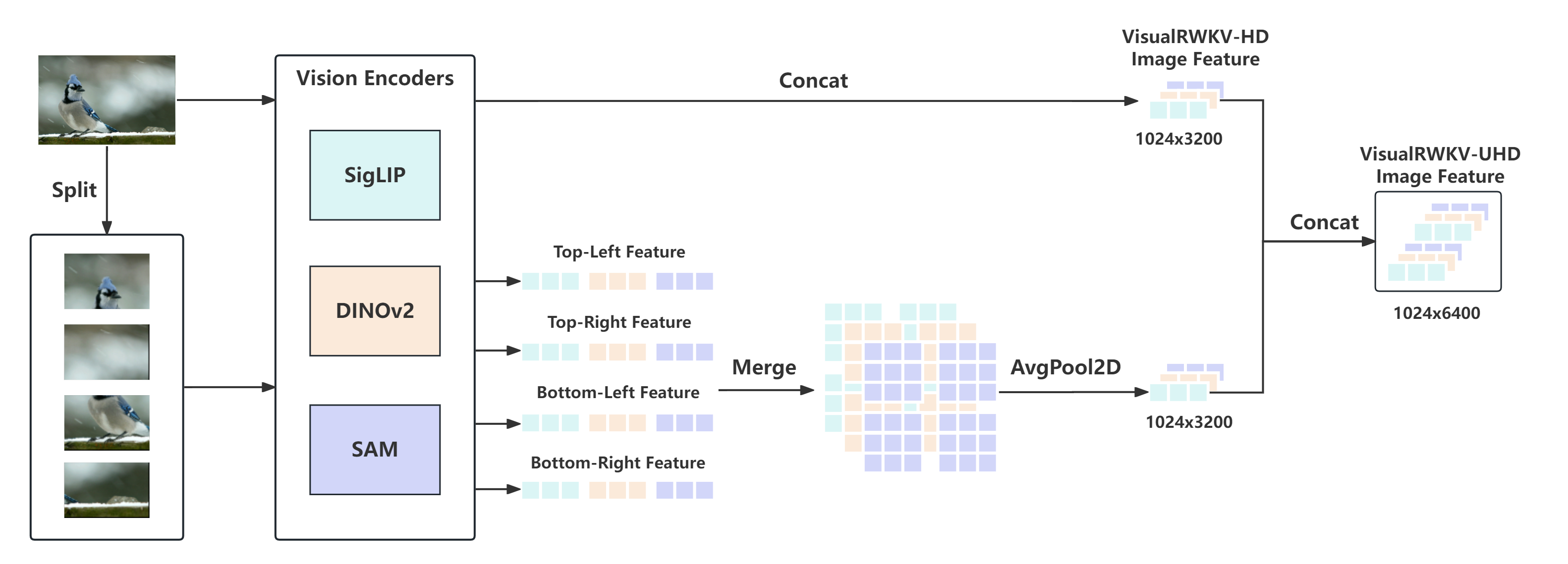}
    \caption{UHD-strategy: We divide the input image into four sections, which are then processed by the SigLip, DINOv2, and SAM encoders. The resulting features from each section are merged and passed through avgpool2d. These pooled features are concatenated with the HD-Feature generated from previous steps, ultimately producing the final UHD-Feature.}
    \label{fig:enter-label}
\end{figure*}

In this model, we further split the image into four parts and then aggregate them, ensuring that the image representation contains both high-resolution and low-resolution features, as shown in Figure 2. This strategy balances coarse features and fine-grained features, increasing the supported resolution to 4096 x 4096. Through this innovative method, the model can more accurately understand and analyze different details of input images when dealing with complex visual information.

\subsection{Projection Layer}

During training, we discovered that the excessive amount of feature information led to adversarial effects within the model. To address this issue, we introduced MLP With Context Gating to replace the traditional linear projection layer. 

Context Gating\cite{miech2017learnable} , a mechanism used to improve neural networks by dynamically adjusting feature representations.. MLP with Context Gating enhances the performance of a multi-layer perceptron (MLP) by adding a gating mechanism that modulates the input features. The Context Gating layer applies a non-linear transformation to the input using a learned weight matrix and a bias term, which are passed through a sigmoid activation. The resulting gate output is multiplied element-wise with the original input, effectively controlling which features are passed through based on the context. 
\begin{equation}
y = \sigma(W_g x + b_g) \odot x
\end{equation}
In this formula, $y$ represents the output, $W_g$ is the gating weight matrix, $b_g$ is the bias term, and $x$ is the input data. The activation function used is the sigmoid function $\sigma$, and the symbol $\odot$ indicates element-wise multiplication. This mechanism optimizes the model's performance by dynamically adjusting the input features.

This adjustment stabilized the training process and significantly improved the model's performance. By managing feature information more effectively, the model could avoid overfitting and instability, thus enhancing overall processing efficiency and accuracy.

\section{Experiments}
In the Experiment section, we evaluate the performance of VisualRWKV-HD and VisualRWKV-UHD models across a variety of tasks, focusing on their ability to handle high-resolution visual inputs effectively. We conducted experiments on several widely-used visual language model (VLM) benchmarks, with a particular emphasis on text-rich and document analysis tasks that benefit from high-resolution image processing.
\begin{table*}[h]
\centering
\resizebox{\textwidth}{!}{ 
\begin{tabular}{llllllllll}
\hline
Model                               & Vision Encoder               & Resolution & SQA   & TextVQA & GQA   & VizWiz & MME            & POPE  & MMB/MMB-CN  \\ \hline
VisualRWKV 1.6B                     & CLIP                         & 336        & 59.05 & 43.57   & 55.23 & 29.84  & 1204.90/245.00 & 0.832 & 55.75/53.17 \\
+ SigLIP and DINOv2                     & SigLIP + DINOv2              & 384        & 53.35 & 41.08   & 56.55 & 31.44  & 1273.67/213.92 & 0.870 & 57.39/51.72 \\
+ SAM-b-1024                    & SigLIP + DINOv2 + SAM-b-1024 & 384        & 57.02 & 48.70   & 58.23 & 30.46  & 1250.50/213.21 & 0.818 & 58.84/57.13 \\
+ Scale up resolution            & SigLIP + DINOv2 + SAM-b-1024 & 448        & 58.55 & 47.75   & \textbf{60.96 }& 33.12  & 1305.38/224.64 & 0.855 & 59.45/53.09 \\
+ MLP with Context Gating        & SigLIP + DINOv2 + SAM-b-1024 & 448        & 54.39 & 54.71   & 60.84 & \textbf{54.97}  & \textbf{1378.62/266.07} &\textbf{ 0.860 }& \textbf{60.31/55.41} \\
+ HD559k dataset & SigLIP + DINOv2 + SAM-b-1024 & 448        &\textbf{ 58.75} & 55.62   & 60.18 & 51.59  & 1271.03/230.36 & 0.857 & 57.56/51.03 \\
+ HD667k dataset & SigLIP + DINOv2 + SAM-b-1024 & 448        & 56.97 & \textbf{56.31}   & 59.52 & 49.88  & 1321.33/232.14 & 0.853 & 58.42/52.84 \\ \hline
\end{tabular}}
\caption{Scaling results of VisualRWKV-HD/UHD models and their performance metrics across different academic tasks.The bolded data in the table represents the best performance. }
\label{tab:visualrwkv_comparison}
\end{table*}

\begin{table*}[h]
\centering
\begin{tabular}{lcccc}
\hline
\textbf{Model}                      & \textbf{Dataset} & \textbf{DocVQA} & \textbf{InfographicVQA} & \textbf{ChartQA} \\ \hline
VisualRWKV 1.6B                     & mix665k          & 10.88           & -                       & 10.00            \\
VisualRWKV 1.6B + MLP               & mix665k          & 11.00           & 11.00                   & 8.00             \\
VisualRWKV 1.6B + MLP (HD)          & HD559k           & 29.11           & 15.53                   & 33.04            \\
VisualRWKV 1.6B + MLP (UHD)         & HD559k           & 35.11           & 16.49                   & 39.32            \\
VisualRWKV 1.6B + MLP (UHD)         & HD667k           & 35.37           & 16.82                   & 40.28            \\ \hline
\end{tabular}
\caption{Scaling results of VisualRWKV-HD/UHD models on Text-rich tasks}
\label{tab:my-table}
\end{table*}

\subsection{Baselines}

In the Baselines section, we compare the performance of the proposed VisualRWKV-HD and VisualRWKV-UHD models with the standard VisualRWKV model to understand the enhancement in model capabilities due to high-resolution processing. The standard VisualRWKV\cite{hou2024visualrwkv} serves as a baseline, representing the foundational architecture of our model family but lacking the high-resolution enhancements. By comparing performance metrics across different versions, we aim to clarify the specific impact of high-resolution processing on text-rich tasks and document analysis.
These tasks often require recognizing small and intricate details, such as fine text and document layouts, making the performance improvements of high-resolution models like VisualRWKV-HD and UHD particularly significant. We compare the results of the baseline model with the high-resolution versions to quantify the benefits of handling higher pixel counts and more detailed inputs, thereby emphasizing the importance of enhancing visual resolution in tasks involving complex visual information.

\subsection{Benchmarks}

We conducted extensive evaluations of VisualRWKV-HD and VisualRWKV-UHD using eight diverse benchmark datasets: SQA\cite{Lu2022LearnTE}, TextVQA\cite{singh2019textvqa}, GQA\cite{Hudson2019GQAAN} (accuracy), VizWiz\cite{bigham2010vizwiz}, MME\cite{fu2023mme}, POPE\cite{li2023pope}, MMB\cite{liu2023mmbench}, and MMB-CN, as shown in Table 1. These datasets encompass a wide range of visual language understanding tasks, each demanding detailed visual and textual interpretation. SQA focuses on scenario-based question answering, requiring the model to comprehend both complex visual and linguistic contexts to provide accurate responses. TextVQA features images embedded with text, challenging the models to accurately extract and interpret textual information from visual inputs. GQA (accuracy) evaluates the models' precision in reasoning over image content, posing questions that demand an in-depth understanding of the image. VizWiz tests the models in a real-world context with noisy and incomplete visual data, a particularly challenging task designed to assist visually impaired individuals.In addition, we incorporated MME for multimodal emotion recognition, where the model is tasked with identifying emotions from both visual and textual inputs. POPE evaluates the model's ability to infer personality traits based on visual content. Lastly, MMB and MMB-CN benchmarks were included to assess the models' performance in multimodal translation and question answering tasks, with MMB-CN specifically designed for Chinese visual language understanding. 
We also conducted tests on document benchmarks, including DocVQA\cite{mathew2021docvqa}, InfographicVQA\cite{mathew2022infographicvqa}, and ChartQA\cite{masry2022chartqa}, to evaluate the performance of our proposed method, as shown in Table 2. DocVQA focuses on document understanding, where our model demonstrated strong performance in interpreting mixed text and image content and effectively answering queries. In InfographicVQA, our method successfully analyzes visually complex infographics, accurately identifying key elements and relationships. Lastly, ChartQA assesses our model's ability to interpret various chart types, revealing excellent performance in understanding chart data and providing insights. Overall, these benchmarks highlight the strengths of our approach in enhancing high-resolution image understanding and its applicability across diverse vision-language tasks.These datasets serve as robust benchmarks to evaluate the models' capabilities in handling high-resolution visual inputs across different modalities and languages. The results from these benchmarks highlight the significant performance improvements of VisualRWKV-HD and VisualRWKV-UHD compared to lower-resolution baselines, demonstrating their effectiveness in processing intricate and high-resolution visual data.

\subsection{Quantitative Evaluation}

In the Quantitative Evaluation, we assess the performance of our proposed VisualRWKV-HD and VisualRWKV-UHD models against several benchmarks to highlight their advancements in handling high-resolution visual inputs, as outlined in the abstract. Table 3 presents a comparative analysis between our models and other leading visual language models (VLMs).
In the SQA benchmark, VisualRWKV-UHD demonstrates its superior ability to process high-resolution images, scoring 56.97, which surpasses Mobile1.7B's 54.7, underscoring the advantage of our model's resolution enhancements. Similarly, in the GQA benchmark, Mobile1.7B scores 56.1, while VisualRWKV-HD and UHD achieve 60.84 and 59.52, respectively, showing a clear improvement due to our innovative lossless downSAMpling and segmented image representation techniques.
For tasks like POPE, where precision is crucial, VisualRWKV-HD outperforms Mobile1.7B with a score of 86.0 versus 84.5, and UHD also exceeds with 85.3. In the MMB benchmark, our models significantly outperform Mobile1.7B, with HD achieving 61.31 and UHD 58.42, compared to Mobile1.7B's 53.2. On the MME Perception task, VisualRWKV-HD excels with a score of 1378.62, noticeably higher than Mobile1.7B's 1196.2, while UHD also performs well at 1321.33.
Interestingly, even in comparison to models with larger parameters, such as Mini-Gemini (2B parameters), VisualRWKV-HD (1.6B parameters) delivers superior results, notably in the MME Perception task (HD: 1378.62, Mini-Gemini: 1341) and MMB (HD: 60.31, Mini-Gemini: 59.8). This underscores the efficiency of our models in visual processing.
Lastly, our models outperform TinyLLaVa-v1 in the GQA benchmark, with VisualRWKV-HD scoring 60.84 and UHD 59.52, while TinyLLaVa-v1 achieves only 57.5.
Overall, the results confirm that the enhancements in visual resolution processing, through lossless downSampling and segmented image techniques, allow VisualRWKV-HD and UHD to excel in high-resolution tasks, providing a clear edge over both lower-resolution models and larger parameter counterparts in complex visual benchmarks.

\begin{table*}[]
\centering
\resizebox{\textwidth}{!}{ 
\begin{tabular}{llcccccccc}
\hline
Method         & LLM              & Resolution & SQA  & TextQA & GQA  & VizWiz & MME      & POPE & MMB/MMB-CN \\ \hline
MobileVLM 1.7B & MobileLLaMA-1.4B & 336        & 54.7 & -      & 56.1 & -      & 1196.2/- & 84.5 & 53.2/-     \\
Mini-Gemini    & Gemma-2B         & 336        & -    & -      & -    & -      & 1341/312 & -    & 59.8/-     \\
TinyLLaVa-v1   & TinyLlama-1.1B   & -          & 59.4 & -      & 57.5 & -      & -        & -    & -          \\
VisualRWKV     & VisualRWKV6-1.6B & 336        & 59.1 & 43.6   & 55.2 & -      & 1204.9/- & 83.2 & 55.8/53.2  \\ \hline
VisualRWKV-HD  & VisualRWKV6-1.6B & 448        & 54.39 & 54.71 & 60.84 & 54.97 & 1378.62/266.07 & 86.0 & 60.31/55.41 \\
VisualRWKV-UHD & VisualRWKV6-1.6B & 448        & 56.97 & 56.31 & 59.52 & 49.88 & 1321.33/232.14 & 85.3 & 58.42/52.84 \\ \hline
\end{tabular}}
\caption{Performance comparison of different visual language models across various academic tasks.}
\label{tab:performance_comparison}
\end{table*}

\subsection{Ablation Study}

In the Experiment section, we assess the performance of the VisualRWKV-HD and VisualRWKV-UHD models across various tasks, emphasizing their effectiveness in handling high-resolution visual inputs. Our experiments target several well-established benchmarks in the realm of visual language models (VLMs), particularly focusing on tasks that are text-rich and involve document analysis, both of which greatly benefit from advanced high-resolution image processing capabilities.
This evaluation aims to illustrate how these models perform in real-world applications where detail and clarity are paramount.

\subsubsection{Ablation on Vision Decoder}
In this section, we compared visual encoders, specifically Siglip and Siglip + DINOv2, based on a resolution of 384, as shown in Table 4. The results showed a comprehensive performance improvement. We further enhanced the model by integrating SAM into the Siglip + DINOv2 framework, leading to additional performance gains on the {SQA}, {TQA}, and MMB/MMB$_{\text{CN}}$
datasets.We assessed the impact of using DINOv2 and SAM on training stability and computational efficiency. The key metrics evaluated included training stability and overall computational cost, as well as the performance results across various datasets.
After introducing DINOv2 and SAM, the model exhibited enhanced stability during training and improved performance across all datasets. This highlights the significant role that the SAM and DINOv2 visual encoders play in effectively processing high-resolution inputs.

\begin{table*}[]
\centering
\resizebox{\textwidth}{!}{ 
\begin{tabular}{llcccccccc}
\hline
Model           & Vision Encoder & Resolution & SQA   & TextVQA & GQA   & VizWiz & MME            & POPE  & MMB/MMB-CN  \\ \hline
VisualRWKV 1.6B & CLIP           & 336        & 59.05 & 43.57   & 55.23 & 29.84  & 1204.90/245.00 & 0.832 & 55.75/53.17 \\
VisualRWKV 1.6B & SigLIP + DINOv2              & 384 & 53.35 & 41.08 & 56.55 & 31.44 & 1273.67/213.92 & 0.870 & 57.39/51.72 \\
VisualRWKV 1.6B & SigLIP + DINOv2 + SAM-b-1024 & 384 & 57.02 & 48.70 & 58.23 & 30.46 & 1250.50/213.21 & 0.818 & 58.84/57.13 \\ \hline
\end{tabular}}
\caption{Ablation study on Vision Encoder
}
\label{tab:visualrwkv_comparison}
\end{table*}

\subsubsection{Ablation on Resolution}
In this experiment, we conducted further research based on the introduction of siglip, DINOv2, and SAM by increasing the resolution from 384 to 448, as shown in Table 5. This adjustment led to improved performance of the model on datasets such as SQA, GQA, and VizWiz.We compared the performance of VisualRWKV-HD and VisualRWKV-UHD under different resolution settings, exploring the impact of increased resolution on accuracy and processing time.
Experimental Results show that: Higher resolutions significantly enhanced accuracy in text-dense tasks. Although inference time increased, techniques such as segmentation and downSAMpling effectively controlled computational costs, achieving a balance between efficiency and accuracy.

\begin{table*}[]
\centering
\resizebox{\textwidth}{!}{ 
\begin{tabular}{llcccccccc}
\hline
Model           & Vision Encoder               & Resolution & SQA   & TextVQA & GQA   & VizWiz & MME            & POPE  & MMB/MMB-CN  \\ \hline
VisualRWKV-HD& SigLIP + DINOv2 + SAM-b-1024 & 384        & 57.02 & 48.70   & 58.23 & 30.46  & 1250.50/213.21 & 0.818 & 58.84/57.13 \\
VisualRWKV-HD& SigLIP + DINOv2 + SAM-b-1024 & 448        & 58.55 & 47.75   & 60.96 & 33.12  & 1305.38/224.64 & 0.855 & 59.45/53.09 \\ \hline
\end{tabular}}
\caption{Ablation study of VisualRWKV-HD on different resolutions.}
\label{tab:visualrwkv_comparison}
\end{table*}

\subsubsection{Ablation on Projection}

In this experiment, we explored the impact of replacing the linear projection method with MLPWithContextGating at a resolution of 448, as shown in Table 6. The goal was to compare models with and without MLPWithContextGating to assess its effect on training stability and computational efficiency.
After incorporating MLPWithContextGating, the model demonstrated significant improvements across various tasks, including TQA, GQA, and POPE. While the degree of enhancement varied, overall performance surpassed that of the model using the linear projection approach.
When MLPWithContextGating was removed, the model's training stability decreased, and memory usage increased, highlighting the importance of this technique in handling high-resolution inputs effectively.These findings suggest that MLPWithContextGating plays a crucial role in enhancing computational efficiency and model stability, particularly when processing high-resolution images.

\begin{table*}[]
\centering
\resizebox{\textwidth}{!}{ 
\begin{tabular}{llcccccccc}
\hline
Model                 & Vision Encoder               & Resolution & SQA   & TextVQA & GQA   & VizWiz & MME            & POPE  & MMB/MMB-CN  \\ \hline
VisualRWKV + Linear Projection& SigLIP + DINOv2 + SAM-b-1024 & 448        & 58.55 & 47.75   & 60.96 & 33.12  & 1305.38/224.64 & 0.855 & 59.45/53.09 \\
VisualRWKV + MLP& SigLIP + DINOv2 + SAM-b-1024 & 448        & 54.39 & 54.71   & 60.84 & 54.97  & 1378.62/266.07 & 0.860 & 60.31/55.41 \\ \hline
\end{tabular}}
\caption{Ablation study on projection layer.}
\label{tab:visualrwkv_comparison}
\end{table*}

\subsubsection{Ablation on Data Scaling up}
In the experiment, we compared the performance of different datasets (mix665k, HD559k, and HD667k) on the VisualRWKV-HD and VisualRWKV-UHD models, focusing on tasks like DocVQA, InfographicVQA, ChartQA, TQA, MME, and VizWiz, as shown in Table 7. The results show that as the dataset size increased, the models' performance improved significantly.
By comparing the use of the mix665k, HD559k, and HD667k datasets, we evaluated their impact on the training stability and computational efficiency of the VisualRWKV-HD and UHD models:
VisualRWKV-HD: After introducing the HD559k dataset, compared to mix665k, the model demonstrated notable improvements in tasks such as DocVQA, InfographicVQA, and ChartQA. The performance in VizWiz and TQA also increased substantially. This highlights how increasing data size and quality enhances the model's ability to handle complex visual language tasks.

VisualRWKV-UHD: With the introduction of the HD559k dataset, the performance in TQA and SQA tasks further improved, confirming the effectiveness of the downSAMpling method in preserving data details while enhancing the model's generalization capabilities. By using larger high-resolution datasets, the model can better understand text-rich visual scenes and demonstrate more robust performance across multiple tasks.
Additionally, in VisualRWKV-UHD, when processing the HD559k and HD667k datasets, dividing the image into four parts and then aggregating them allows the model to integrate both high-resolution and low-resolution features. This approach strengthens image representation, enabling the model to perform better on tasks requiring detailed visual analysis.

\begin{table*}[ht!]
\centering
\resizebox{\textwidth}{!}{ 
\begin{tabular}{lllcccccccc}
\hline
Model                          &  Dataset&Vision Encoder               & Resolution & SQA   & TextVQA & GQA   & VizWiz & MME            & POPE  & MMB/MMB-CN  \\ \hline
VisualRWKV-UHD&  mix665k&SigLIP + DINOv2 + SAM-b-1024 & 448 & 54.39 & 54.71 & 60.84 & 54.97 & 1378.62/266.07 & 0.860 & 60.31/55.41 \\
VisualRWKV-UHD&  HD559k&SigLIP + DINOv2 + SAM-b-1024 & 448        & 58.75 & 55.62   & 60.18 & 51.59  & 1271.03/230.36 & 0.857 & 57.56/51.03 \\
VisualRWKV-UHD&  HD667k&SigLIP + DINOv2 + SAM-b-1024 & 448        & 56.97 & 56.31   & 59.52 & 49.88  & 1321.33/232.14 & 0.853 & 58.42/52.84 \\ \hline
\end{tabular}
}
\caption{Ablation study of VisualRWKV-UHD on different datasets.}
\label{tab:visualrwkv_comparison}
\end{table*}

\subsection{Efficiency Analysis}

The method proposed in this paper primarily concatenates features along the channel dimension, maintaining a cap of 1024 image tokens without increasing their number. Unlike other methods that inflate the input image tokens and consequently impact the inference speed of vision-language models (VLMs), our approach preserves efficiency.

During the prefill stage, VisualRWKV-UHD processes four times more images than VisualRWKV-HD, which results in slightly lower efficiency but enhances the understanding of high-resolution images. In the decode stage, however, the speeds of VisualRWKV-UHD and VisualRWKV-HD are nearly identical. This enables us to fully leverage the high-efficiency inference capabilities of RWKV, resulting in faster inference speeds and reduced memory usage compared to Transformer-based VLMs.
\section{Conclusion}
In this paper, we presented VisualRWKV-HD and VisualRWKV-UHD, two advanced models in the VisualRWKV family, designed to excel in processing high-resolution visual inputs. Through comprehensive evaluations across a range of benchmarks, including SQA, GQA, and VizWiz, we demonstrated that these models significantly outperform traditional approaches, particularly in text-rich and document analysis tasks.

The introduction of techniques such as lossless downsampling, segmented image representation, and the integration of robust visual encoders (DINOv2 and SAM) has not only enhanced the accuracy and efficiency of our models but also stabilized training processes. Comparative analysis with LLaVA-UHD revealed that VisualRWKV models achieve a better balance of computational cost, memory efficiency, and processing speed, making them suitable for real-time applications that demand high precision.

The findings highlight the importance of high-resolution processing in complex visual language tasks, underscoring the potential of VisualRWKV-HD and UHD to serve as valuable tools in real-world applications requiring detailed visual interpretation. Our results advocate for continued research into optimizing model architectures and techniques to further enhance visual processing capabilities in visual language models.

\textbf{Limitations}

Despite the significant performance improvements of VisualRWKV-HD and VisualRWKV-UHD models, several limitations persist. Firstly, their high computational and memory requirements may restrict accessibility for users with less powerful hardware, particularly affecting real-time applications. Additionally, the models rely heavily on the quality and quantity of training data; limited availability of high-quality labeled data could compromise their effectiveness. The lack of interpretability in decision-making processes remains a challenge, especially in critical fields like healthcare and finance. Furthermore, while the models excel in benchmark tasks, their generalization capabilities in diverse and unfamiliar scenarios need further validation. Finally, the integration of these models with other modalities, such as audio or tactile data, is an area yet to be fully explored, which could limit their applicability in multi-modal learning contexts. Addressing these limitations is essential for enhancing the practical usability of VisualRWKV models in real-world applications.

\bibliography{custom}

\appendix
\newpage
\onecolumn

\appendix

\section{Model Architecture and Computing}

\textbf{Model Architecture:} The VisualRWKV models used in our experiments are visual extensions of the Recurrent Weighted Key-Value (RWKV) architecture, designed to handle both visual and textual data. We experimented with the following configurations:
\begin{itemize}
    \item \textbf{VisualRWKV 1.6B:} A baseline model using 1.6 billion parameters.
    \item \textbf{VisualRWKV 1.6B + MLP:} Enhanced with a Multi-Layer Perceptron (MLP) to improve feature extraction.
    \item \textbf{VisualRWKV 1.6B + MLP (HD/UHD):} Models utilizing High Definition (HD) and Ultra High Definition (UHD) strategies for fine-grained feature extraction.
\end{itemize}

\textbf{Computing Infrastructure :} Infrastructure A range of computational resources were employed in the study. The standard training and benchmark evaluation were conducted using 8 NVIDIA A100-80GB GPUs. The VisualRWKV 7B model is trained with 6 A100 GPUs due to insufficient memory capacity with 8 GPUs. For the efficiency analysis, we employed an NVIDIA RTX 3090 GPU.

\textbf{Computing Budget:} Training an epoch of VisualRWKV 1.6B with 8 A100 GPUs takes 6.7 hours, equivalent to 53.6 GPU hours; Training an epoch of VisualRWKV 3B with 8 A100 GPUs takes 11.3 hours, equivalent to 90.4 GPU hours; Training an epoch of VisualRWKV 7B with 6 A100 GPUs takes 26.5 hours, equivalent to 159 GPU hours 

In all cases, the RWKV backbone was adapted for visual tasks by incorporating Vision Encoders and using Context Gating. These models were fine-tuned for visual question-answering tasks on various datasets.

\section{Datasets}

We trained and evaluated the models on the following datasets:

\begin{itemize}
    \item \textbf{mix665k:} This is the dataset used by LLaVA for instruction tuning, comprising 665,000 diverse images aimed at enhancing the model's adaptability to various visual tasks and instructions, thereby improving its overall performance and usability.
    
    \item \textbf{HD559k:} This dataset is our custom high-resolution dataset consisting of 559,000 high-quality images. It focuses on testing the model's performance when processing high-quality visual content, particularly in terms of detail, color, and clarity, ensuring that the model can accurately capture complex visual information. Table 8 and Figure 3 provide an overview of the data proportions in the HD559k dataset.
    
    \item \textbf{HD667k:} As another significant contribution from our team, HD667k is a larger high-resolution dataset containing 667,000 images. This dataset not only enriches the training data for the model but also provides additional support for its performance in diverse and complex visual scenarios, helping to improve the model's generalization ability and robustness in practical applications. Table 8 and Figure 3 provide an overview of the data proportions in the HD667k dataset.
\end{itemize}
\begin{table}[h]
    \centering
    \begin{tabular}{|l|c|}  
        \hline
        \textbf{Dataset Name} & \textbf{Quantity} \\
        \hline
        textocr & 21.9k \\
        DocReason25K & 25k \\
        sharegpt4v\_instruct\_61k & 61k \\
        monkey\_685k\_multi\_round & 294k \\
        llavar\_16k & 16k \\
        pdfa-eng-50k & 50k \\
        pdfa-eng-9k-multi\_sft & 9k \\
        idl\_train-35k & 35k \\
        cord-v2-fix2 & 0.8k \\
        llava\_mix50k & 50k \\
        \hline
    \end{tabular}
    \caption{Overview of Datasets Used of HD559k}
    \label{tab:datasets}
\end{table}

\begin{figure}
    \centering
    \includegraphics[width=1\linewidth]{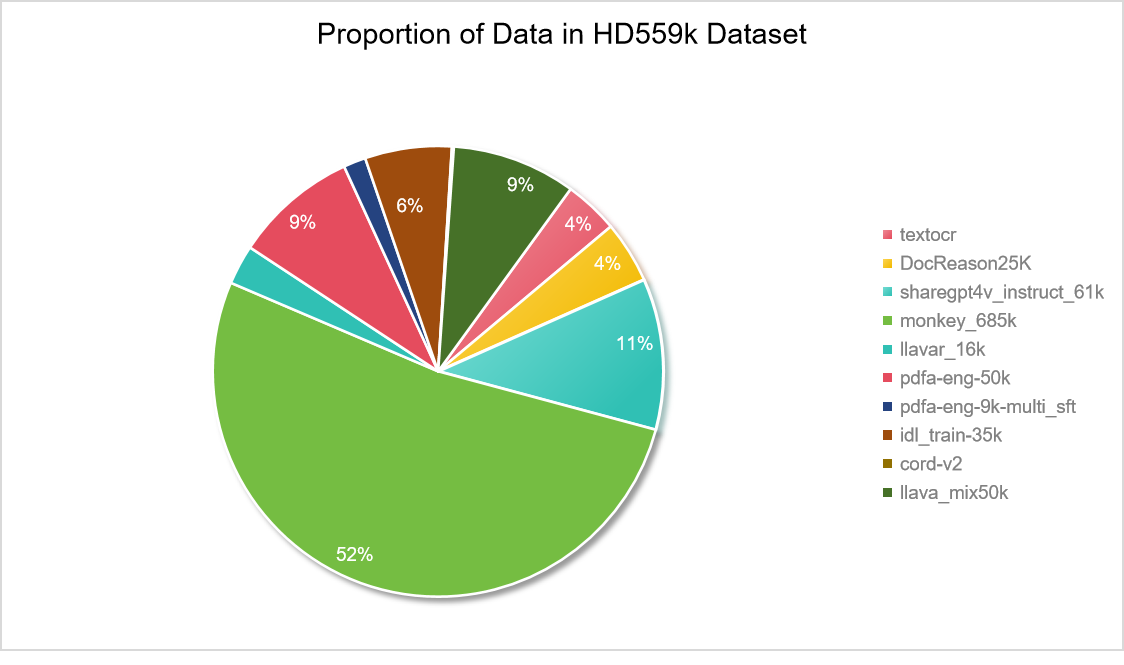}
    \caption{Distribution of the HD559k dataset, showcasing the various datasets and their respective quantities. This comprehensive dataset includes a diverse range of sources, contributing to a total of 559,494 images utilized for training and evaluation purposes.}
    \label{fig:enter-label}
\end{figure}

\begin{table}[h]
    \centering
    \begin{tabular}{|l|c|}  
        \hline
        \textbf{Dataset Name} & \textbf{Quantity} \\
        \hline
        textocr & 21.9k \\
        DocReason25K & 25k \\
        sharegpt4v\_instruct\_61k & 61k \\
        monkey\_685k\_multi\_round & 294k \\
        llavar\_16k & 16k \\
        pdfa-eng-50k & 50k \\
        pdfa-eng-9k-multi\_sft & 9k \\
        idl\_train-35k & 35k \\
        cord-v2-fix2 & 0.8k \\
        llava\_mix50k & 50k \\
        chart2text & 26.9k \\
        rendered\_text & 10k \\
        iam & 5.66k \\
        st\_vqa & 17.2k \\
        tabmwp & 22.7k \\
        vistext & 9.97k \\
        visualmrc & 3k \\
        websight & 10k \\
        infographic\_vqa & 2.1k \\
        \hline
    \end{tabular}
    \caption{Overview of Datasets Used of HD667k}
    \label{tab:datasets}
\end{table}

\begin{figure}
    \centering
    \includegraphics[width=1\linewidth]{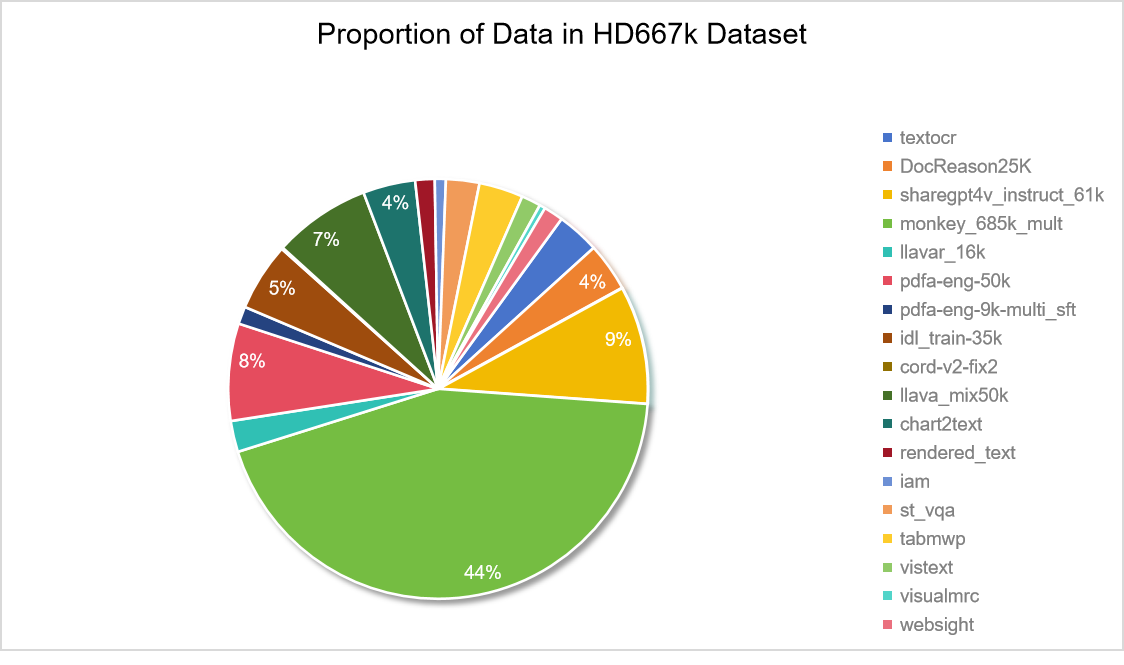}
    \caption{Distribution of the HD667k dataset, illustrating the composition and quantity of various datasets included. With a total of 667,000 images, this dataset encompasses a wide array of visual tasks and sources, aimed at enhancing the training and evaluation of model performance.}
    \label{fig:enter-label}
\end{figure}

\section{Experimental Setup}

\textbf{Preprocessing:} Input images were divided into four sections, each encoded by three vision encoders (SigLIP, DINOv2, SAM). Features were merged and passed through an MLP with Context Gating.

\textbf{Training:} The models were trained using the AdamW optimizer with a learning rate of \(X\), using NVIDIA GPUs and mixed precision. Training continued for 100 epochs, with early stopping applied after 10 epochs of no improvement.

\textbf{Evaluation:} Models were evaluated on the DocVQA, InfographicVQA, and ChartQA datasets. These datasets represent different challenges, from document understanding to infographics and chart analysis.

\section{Data and Hyperparameters}
\begin{itemize}

\item \textbf{A. Training Data}

We used a two-phase training process for VisualRWKV. In the \textit{Feature Alignment Phase}, 558K images from LAION-CC-SBU were utilized to connect a frozen vision encoder with a frozen LLM. This phase establishes the foundation for robust image-text alignment. In the \textit{Visual Instruction Tuning Phase}, an expanded dataset of 150K multimodal examples generated by GPT and 515K VQA datasets were used to enhance the model’s capacity for multimodal tasks.All the data used in this paper are consistent with their intended use. 

Ethical guidelines were strictly followed in data preparation, focusing on identifying and handling PII and sensitive content via automated tools and manual reviews. Anonymization techniques, such as data masking, were applied to ensure data integrity and privacy.

\item \textbf{B. Evaluation Benchmarks}

We employed various benchmarks to evaluate the model. VQA-v2 and GQA metrics are based on the test-dev split, while TextVQA is evaluated on its validation set. ScienceQA and POPE metrics are from their respective test sets. MMBench metrics are based on the development set, and MME is evaluated on a specific test set.

\item \textbf{C. Data Language}

Our training data spans multiple datasets, with most Visual Question Answering (VQA) datasets being in English. The ShareGPT data is multilingual, covering multiple user-contributed languages. Among the evaluation benchmarks, MMBench-cn is in Chinese, while the rest are in English. 

\item \textbf{D. Hyperparameters}

The models used 1.6B parameters for experiments. Detailed hyperparameters for both the vision-language alignment pretraining and the visual instruction tuning phases are listed in Table 10. These include settings optimized for diverse tasks across different datasets, ensuring robust model performance.
\begin{table}[ht]
\centering
\small
\begin{tabular}{|l|c|c|}
\hline
\textbf{Hyperparameter} & \textbf{1.6B-Pretrain} & \textbf{1.6B-Finetune} \\ \hline
batch size               & 256                    & 128                     \\ 
lr init                  & 1e-3                   & 6e-5                    \\ 
lr end                   & 1e-5                   & 1.5e-5                  \\ 
lr schedule              & cosine decay           & cosine decay            \\ 
lr warmup ratio          & 0                      & 0                       \\ 
weight decay             & 0                      & 0                       \\ 
epoch                   & 2                      & 2                       \\ 
optimizer                & AdamW                  & AdamW                   \\ 
DeepSpeed stage          & 1                      & 1                       \\ \hline
\end{tabular}
\caption{Hyperparameters for 1.6B model pretraining and finetuning.}
\label{tab:hyperparameters}
\end{table}
\end{itemize}
\section{Limitations and Future Work}

Although the UHD strategies significantly improve model performance, especially on ChartQA, challenges remain in document understanding tasks. Future work will explore improved feature extraction methods and further optimize the model for multimodal tasks.

\section{Use of AI Assistants}
 In this research, an AI writing assistant is solely employed for the purposes of paraphrasing, spell-checking,
 and enhancing the author’s original content, and it does not introduce any novel content.

\end{document}